\documentclass[runningheads]{llncs}
\usepackage{graphicx}
\usepackage{booktabs}
\usepackage{cite}
\usepackage{amsmath,amssymb,amsfonts}
\usepackage{algorithmic}
\usepackage{textcomp}
\usepackage{xcolor}
\usepackage{booktabs}

\begin{document}

\title{Investigating the role of educational robotics in formal mathematics education: the case of geometry for 15-year-old students\thanks{Supported by the NCCR Robotics, Switzerland}
}

\titlerunning{Investigating the role of ER in formal mathematics education}
\authorrunning{Brender and El-Hamamsy, et al.}

\author{Jérôme Brender\inst{1,2,} \thanks{Both authors contributed equally to this work}, 
Laila El-Hamamsy\inst{1,2,\star\star}\orcidID{0000-0002-6046-4822}, 
Barbara Bruno\inst{2,3}\orcidID{0000-0003-0953-7173}, 
Frédérique Chessel-Lazzarotto\inst{1}, 
Jessica Dehler Zufferey\inst{1}\orcidID{0000-0001-5163-807X}, 
Francesco Mondada\inst{1,2}\orcidID{0000-0001-8641-8704}
}

\institute{Center LEARN, Ecole Polytechnique Fédérale de Lausanne (EPFL), Switzerland \and 
Mobots Group, Ecole Polytechnique Fédérale de Lausanne (EPFL), Switzerland \and
CHILI Lab, Ecole Polytechnique Fédérale de Lausanne (EPFL), Switzerland
\\
\email{firstname.lastname@epfl.ch}}

\maketitle
\begin{abstract}

Research has shown that Educational Robotics (ER) enhances student performance, interest, engagement and collaboration.
However, until now, the adoption of robotics in formal education has remained relatively scarce.
Among other causes, this is due to the difficulty of determining the alignment of educational robotic learning activities with the learning outcomes envisioned by the curriculum, as well as their integration with traditional, non-robotics learning activities that are well established in teachers' practices. 
This work investigates the integration of ER into formal mathematics education, through a quasi-experimental study employing the Thymio robot and Scratch programming to teach geometry to two classes of 15-year-old students, for a total of 26 participants.
Three research questions were addressed: (1) Should an ER-based theoretical lecture precede, succeed or replace a traditional theoretical lecture? (2) What is the students’ perception of and engagement in the ER-based lecture and exercises? (3) Do the findings differ according to students’ prior appreciation of mathematics?
The results suggest that ER activities are as valid as traditional ones in helping students grasp the relevant theoretical concepts. Robotics activities seem particularly beneficial during exercise sessions: students freely chose to do exercises that included the robot, rated them as significantly more interesting and useful than their traditional counterparts, and expressed their interest in introducing ER in other mathematics lectures. Finally, results were generally consistent between the students that like and did not like mathematics, suggesting the use of robotics as a means to broaden the number of students engaged in the discipline.

\end{abstract}




\section{Introduction} \label{introduction}

Research has shown that Educational Robotics (ER) can be used as a tool to enhance teaching \cite{alimisis2013educational} and learning \cite{miller2016robotics}, from early childhood \cite{bers_teaching_2005} to tertiary education \cite{benitti_how_2017}. ER provides ``an experimental platform for practice'' \cite{karim2015review} and improves students' motivation, engagement and learning \cite{Rogers2004BringingET}. 
While ER can be viewed as a tool fitting many and varied disciplines \cite{jung_systematic_2018}, it is most commonly associated with Computer Science \cite{el-hamamsy_computer_2021} and STEM related disciplines \cite{karim2015review, benitti_how_2017} such as mathematics \cite{zhong_systematic_2020}.
Papert, who instigated the learning theories on constructionism\footnote{The constructionist theory of learning stipulates that knowledge is built more effectively when people are actively engaged in building tangible and shareable artefacts.} \cite{papert_mindstorms:_1980}, was one of the first to employ ER for mathematics, in the 80's. He used the LOGO programming language to teach geometry and found that ER improves children's motivation, learning and interaction in the classroom \cite{papert_mindstorms:_1980}. 
Given such premises, it is thus surprising to find that only a limited number of studies explore the benefits of introducing robotics into formal mathematics education \cite{silk2010designing, ollero_methodology_2018}. 
In 2019, Leoste and Heidmets \cite{leoste_impact_2019} conducted a longitudinal study with students from 20 classes, confirming that the use of ER in mathematics lessons improved students' learning outcomes on a national standardised assessment.
Their results are coherent with recent studies on the use of the ``Concreteness Fading'' method in mathematics: Kim \cite{Hee_2020} found that starting with physical activities that include manipulatives (such as ER) and ``gradually fading concreteness to access abstract concepts’’ effectively supports ``students [access to] conceptual understanding in mathematics classrooms’’. These findings suggest that ER could play a pivotal role in a ``Concreteness Fading Strategy'' to improve learning outcomes in the formal mathematics curriculum. 

Despite its numerous benefits, the use of robots in formal education settings, as opposed to extra-curricular activities, is still relatively sparse \cite{benitti_how_2017}, both in terms of research and practice. A recent review by Zhong and Xia \cite{zhong_systematic_2020} on the use of educational robots in mathematics education concluded that more research was required ``to further explore the integration of robotics and mathematics education’’.
Progressing in such research requires facilitating the introduction of ER into regular classrooms.
Unfortunately, teachers, who play a determining role in the classroom, are often preoccupied by time \cite{chevalier_pedagogical_2016, el-hamamsy_computer_2021} and need to be assured that the use of Educational Robots will help reach the learning outcomes without incurring in a loss of time. Furthermore, the research is well aware of the importance of providing teachers with adequate guidelines for activity design \cite{Giang_281683} and intervention \cite{chevalier_fostering_2020}, to support the alignment of ER learning activities with the learning outcomes of the curriculum. However the reality is often far from these principles, leaving teachers to face the difficulties of integrating ER activities in their practices alone \cite{Eguchi}. 

In an effort to contribute to the study of effects and modalities of the integration of ER in mathematics formal education, in this article we specifically address the following research questions: \emph{1) Should an ER-based theoretical lecture precede, succeed, or replace a traditional theoretical lecture? 2) What is the students’ perception of, and engagement in, the ER-based lecture and exercises? 3) Do the findings differ according to students’ prior appreciation for mathematics? 
I.e., can ER help broaden the number of students successfully engaged in the discipline?} The methodology devised to investigate the afore-listed questions (see section \ref{sec:methodology}), and the results herein reported (see section \ref{sec:results}), constitute the main contributions of this article. 
Additionally we provide in open-source all the ER-based pedagogical resources devised\footnote{For the open-source ER-based pedagogical resources devised see here \url{10.5281/zenodo.4649842}} 
and used in this study\footnote{The proposed ER learning activities follow the structure outlined for 11th grade (15 y.o. students) mathematics in the mandatory curriculum of the 
Canton Vaud, Switzerland.}.

\section{Methodology}
\label{sec:methodology}

A key requirement for the assessment of ER learning activities in the context of formal mathematics education is the presence of adequate ER content. Section \ref{Design_of_the_educational_robotics_mathematics_unit} outlines the learning unit considered for this study and describes the proposed ER-based theoretical and exercise activities for a geometry lecture. Details about the study participants and experimental design are provided in Section \ref{Participants_and_Data_Collection}.

\subsection{Design of the ER content}\label{Design_of_the_educational_robotics_mathematics_unit}

In our study, we consider the case of the formal mathematics curriculum at the level of secondary school in Switzerland. 
A common practice for Swiss mathematics teachers is to start with the theoretical introduction of a concept and then proceed to paper-based exercises. Students are often free to choose the order in which to do the exercises. 
Based on this pedagogical approach, and existing material, we thus designed an ER-based theoretical introduction lecture for the curriculum topic of ``planar geometric figures'' and a set of related ER-based exercises. 

\subsubsection{Choice of the ER platform}

All the learning activities we designed rely on the Thymio II robot \cite{mondada_bringing_2017} (henceforth referred to as Thymio), which was chosen because (1) its structure and shape is well suited for attaching a pencil and making it draw geometric figures; and (2) it is presently being introduced into classrooms in the region \cite{el-hamamsy_computer_2021} and thus already familiar for a number of teachers. 
Moreover, Thymio has been successfully employed as an educational tool in a variety of settings, ranging from primary school \cite{el-hamamsy_computer_2021} to university\footnote{
\url{https://edu.epfl.ch/coursebook/en/basics-of-mobile-robotics-MICRO-452}}.
This versatility is rendered possible by the spectrum of programming languages that can be used with it (including three block-based visual programming languages: Blockly \cite{mondada_bringing_2017}, VPL \cite{mondada_bringing_2017}, Scratch \cite{resnick2009scratch}; and a text based programming language, ASEBA \cite{mondada_bringing_2017}). 
Scratch was selected for the present context as it is particularly adapted for our target group (15-year-old students) \cite{foerster_integrating_2016} and has already been used in various studies to teach mathematics \cite{foerster_integrating_2016,Iskrenovic_Momcilovic_Olivera}. 

\subsubsection{ER-based theoretical lecture} \label{Theoretical_lectures_including_robotics}

The ER lecture was designed starting from the guidelines for the theoretical introduction of planar geometric figures provided in the official regional study plan\footnote{The study plan and learning outcomes outlined for students throughout compulsory education in the french speaking region of Switzerland is publicly accessible at \url{https://www.plandetudes.ch}
}, and adhering as closely as possible to the way the teachers in the study introduce new mathematics topics. 
In the designed ER lecture, the teacher programs Thymio to illustrate the various concepts students are expected to acquire. 
Specifically, as the students are expected to understand what a regular polygon is, know its properties and how to construct it, the teacher illustrates the construction of complex polygons (e.g. hexagons, octagons...) by programming Thymio and making it draw polygons on a sheet of paper.
Additionally, as the students are also expected to recognise and name the various angles in parallel lines\footnote{Facing angles - or vertically opposite angles, corresponding angles, complementary angles and alternate exterior/interior angles.}, two Thymios are concurrently used to demonstrate the relationships between these angles. 
The material prepared for the lecture is available at \url{10.5281/zenodo.4649842}. 
The lecture was designed to last 90 minutes, which equals the duration of the corresponding traditional theoretical lecture.

\subsubsection{ER-based exercises}\label{Exercise_sessions_including_robotics}

Starting from the 31 exercises present in the curriculum, we designed 6 robot-based activities which, while functionally equivalent to their pen-and-paper counterparts, are centred on the programming of, and/or interaction with, the Thymio robot for their resolution.
As commonly done for robotics activities, our ER exercises envision that students work in pairs, a setup known to foster collaboration and often preferred to individual settings \cite{Pair_programming}.
To detach the proposed ER exercises from programming, students were provided with pre-filled code snippets, and thus required only the basic programming skills that were covered during the ER introduction. 
Once a code snippet is loaded for execution on the robot (i.e., assumed by students to be the right answer), its correctness is immediately and directly assessed by observing the figure drawn by the robot. This is an interesting feature of ER activities, as the benefits of feedback mechanisms allowing learners to verify the correctness of their solutions have been shown in previous studies \cite{Azevedo_1995} and are typically lacking in traditional pen-and-paper exercises.

\subsubsection{Validation of the ER content}

The designed ER lecture and exercises were submitted to 5 teachers with diverse background and experience to ensure their alignment with learning objectives and methods.
Specifically, our experts included one pre-service teacher, three experienced in-service mathematics teachers and a teacher who transitioned to research on digital education.

\subsection{Participants and Study Design} \label{Participants_and_Data_Collection}
\begin{figure}[h]
  \centering
  \vspace{-5pt}
  \includegraphics[width=0.8\linewidth]{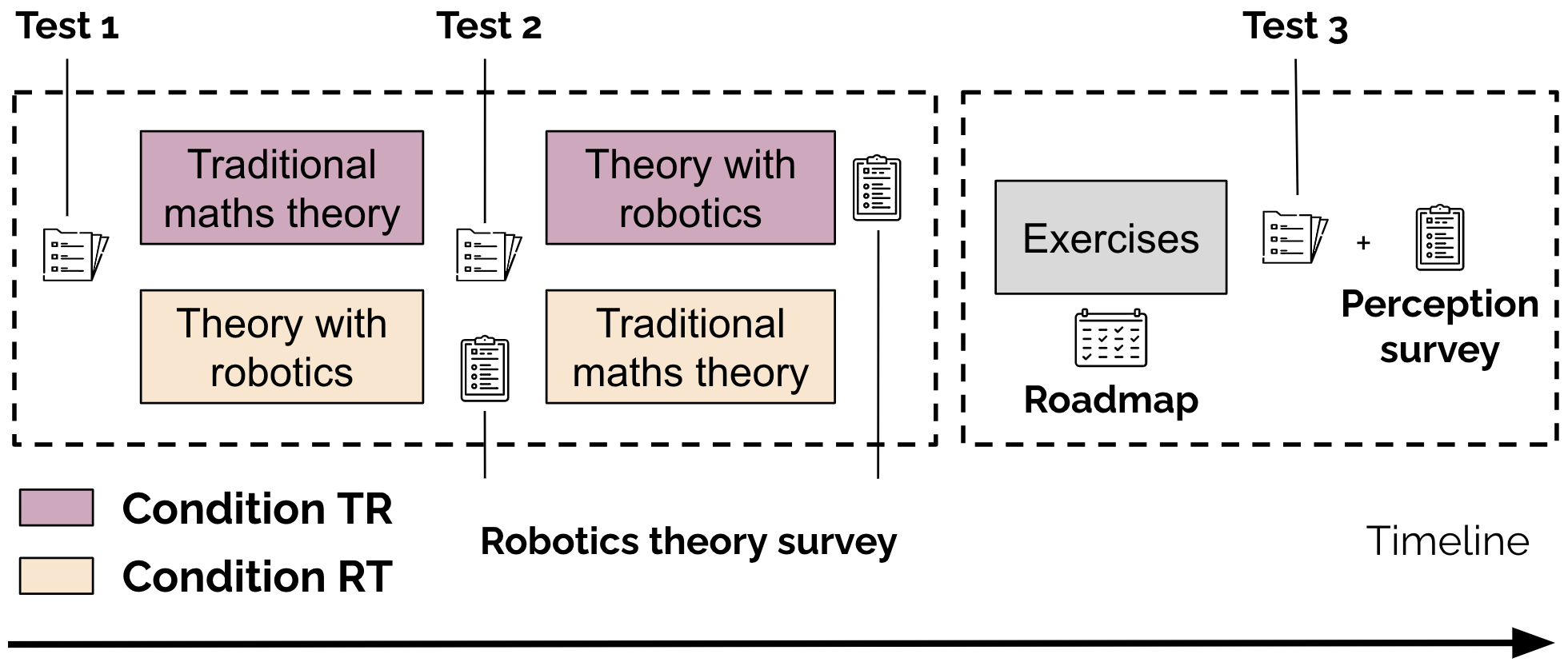}
  \vspace{-10pt}
  \caption{Study design with each class being split in two to have half the students of each class in Condition TR and the other half in RT.}
  \label{fig:scheduled}
\end{figure}

The study was conducted in a public school in Switzerland, with two classes of grade 11 students (15 y.o.) taught by two different mathematics teachers. A total of 26 students participated in the study (16 boys and 10 girls), most of which had no prior experience in robotics, CS or Scratch-based programming. 
The overall outline of the study is reported in Fig. \ref{fig:scheduled}, with each class being split in two to avoid a confound between the teacher and the order effect. As anticipated, both the traditional and the ER-based theoretical lecture lasted 90 minutes, while the exercise session (in which students are presented with traditional and ER-based exercises) lasted approximately 225 minutes (5 teaching periods), corresponding to the time allocated to exercises in the official study plan.
Details on the assessment tools used during the study are reported in Table \ref{tab:all_assessment}.
The surveys administered focused mainly on interest \cite{ryan2000self} and perceived utility \cite{King_2006} to gain insight into the students' intrinsic motivation \cite{ryan2000self} to introduce ER into mathematics formal education. Each question is administered on a 7-point Likert scale (score between -3 and 3, 0 being neutral). 

\begin{table*}[ht]
\caption{Summary of the data collected during the study (tests, surveys and roadmap)}
\vspace{-5pt}
\label{tab:all_assessment}
\ifdefined\USEACM
\begin{tabular}{p{2cm}p{4cm}p{7.85cm}l}
\else 
\begin{tabular}{p{1.5cm}p{2cm}p{6.75cm}p{1.25cm}}
\fi
\toprule
\ifdefined\USEACM
Assessment &
\else 
Data &
\fi
  Objective &
  Content &
  Answer format \\ \midrule
Test 1 &
  Reference performance &
  3 exercises on prior geometry knowledge
  &
  Paper-based\\
Test 2 &
  Performance after the first lecture &
  2 exercises of prior geometry knowledge, 2 on the first theme (polygons),   2 on the second (angles)  &
  Paper-based \\
Test 3 &
  End of unit performance&
  Similar to Test 2  & Paper-based \\ 
Robotics theory survey &
  ER-based theory motivation &
  I found the theory with robotics 1) interesting; 2) useful & 7-point Likert
  \\ \midrule
Roadmap &
  ER-based &
  Order of exercise completion &
  Integer \\ 
  & exercises engagement &  Activity type (ER-based done with the robot, ER-based done without the robot, Traditional) & Checkbox \\ 
  & &
  The exercises were 1) interesting; 2) useful &
  7-point Likert \\ 
  Perception survey & Perception of ER-based content &
  1) Interest (I enjoyed doing them, was interested), 2) Collaboration (I discussed with my classmate, collaborated with my classmate to find the answers), 3) Ease (I found the activities easy, am sure of my answers, did well), 4) Effort (I was concentrated, did the activities as well as possible), 5) Future interest (I would like to do similar activities in maths, in other disciplines, would recommend such exercises to others for maths). &
  7-point Likert \\
  & Maths appreciation & I generally like mathematics &
  7-point Likert \\ \bottomrule
\end{tabular}
\vspace{-10pt}
\end{table*}

\subsubsection{RQ1 - Should an ER-based theoretical lecture precede, succeed or replace a traditional theoretical lecture?}\label{RQ1}

Our hypothesis is that starting with a concrete experience (i.e. with the ER-based lecture) and moving to a theoretical lecture helps the students have a better ``conceptual understanding'' of geometry. To verify this hypothesis, we designed a between-subjects experiment comparing the condition \textit{Traditional-Robotic (TR)} (i.e., starting with the traditional lecture and moving on to the ER-based lecture, shown in purple in Fig. \ref{fig:scheduled}) with the condition \textit{Robotic-Traditional (RT)} (i.e., starting with the ER-based lecture and moving on to the traditional lecture, shown in orange in Fig. \ref{fig:scheduled}).
Differences between the two conditions in terms of academic performance are assessed at three points in time (see Fig. \ref{fig:scheduled}). Test 1, administered prior to the start of the experiment, assesses students' knowledge in the geometry concepts identified as pre-requisite for the considered unit. Test 2, administered at the end of the first theoretical lecture, assesses students' understanding of the presented content, and is similar to test 3, administered at the end of the exercise sessions. All tests are based on the assessments in the official curriculum. 
RQ1 was therefore evaluated by checking the learning gains computed for Test 2 and Test 3 with respect to the baseline provided by Test 1, between the two experimental arms.
To mitigate a possible teacher effect, half of each class was placed in condition RT and the other half in TR, with the responsible teacher giving the traditional lecture and a researcher (the same for both classes) giving the ER-based one. Students were assigned to conditions to ensure they had similar distributions of competency in the discipline based on their performance in the course, as assessed by their teachers.

\subsubsection{RQ2 - What is the students’ perception of, and engagement in, the ER-based lecture and exercises?}

Our hypothesis is that students would find the ER-based lecture and exercises interesting and useful, manifesting in high engagement in these activities and interest to integrate robotics into other mathematics lessons in the long term \cite{leoste_impact_2019}. 
Students' appreciation for the proposed ER-based activities was assessed with a within-subjects experiment and via two complementary approaches: an objective assessment of their behaviour during the exercise session and a subjective assessment of their perception of the ER-based theoretical lecture and exercises. 
Throughout the exercise sessions students are free to decide which (among the 31 traditional and 6 ER-based) exercises to address, in which order and in which manner. Students could in fact pick a traditional exercise and solve it without using the robot (referred to as ``traditional''); an ER-based exercise and solve it using the robot to validate their solutions (referred to as ``robotics with robot''); or, lastly, the same ER-based exercises that the students decided to solve without using the robot (referred to as ``robotics without robot'').
In the \textit{Roadmap}, students were thus asked to report the order in which they did the exercises and, for ER-based ones, whether they used the robot or not. 
Students' appreciation for the ER-based theoretical lecture and exercises was measured in terms of perceived interest and usefulness, respectively with the \textit{Robotics theory survey} and the \textit{Roadmap}, which also allows for the analysis of students' behaviour. 
At the end of the experiment, students were administered a final \textit{Perception} survey, to evaluate their perception of the proposed ER-based content from the perspectives of interest, collaboration, facility (with respect to solving the exercises), effort and future interest (with respect to including robotics in future mathematics lessons, as well as lessons of other disciplines). Each of these items corresponded to a minimum of 2 questions, to acquire a more reliable estimate of the construct from the students \cite{fowler1995improving}. Internal consistency is calculated using Cronbach's alpha \cite{cronbach2004my}.

\subsubsection{RQ3 - Do the findings differ according to students’ prior appreciation for mathematics?}
Our hypothesis is that ER would interest students and help engage those that are generally less invested in the mathematics curriculum, and thus compensate for differences in terms of prior mathematics appreciation.
To investigate this question, we rely on the data collected for RQ1 and RQ2, categorising students based on their reported liking of mathematics on a 7-point Likert scale (13 do, 7 don’t). More specifically, we consider differences in performance (learning gain), the level of engagement in the exercise sessions and the overall perception of the robotics-enhanced geometry lecture.

\section{Results}
\label{sec:results}

\subsection{RQ1 - Lecture type and students' learning}

Test 1 was administered prior to the start of the interventions and was used to ensure that the students in both conditions had similar levels of prior knowledge (Kruskal Wallis test fails to reject $H_0$, $p>0.05$).
The students' performance in Test 2 and Test 3 exceeded 80\% for both experimental arms (Test 2: $84\pm11$\% for condition RT, and $84\pm 15$\% for condition TR; Test 3:  $86\pm8 $\% for condition RT and $87\pm11$\% for condition TR). No significant difference was found between the two conditions (Kruskal Wallis test fails to reject $H_0$, $p>0.05$). The finding thus seems to suggest that the ER-based lecture is equivalent to the traditional one in terms of students' learning, with both groups being sufficiently prepared to move on to the exercises after their first theory lecture. The lack of significant progress from test 2 to test 3 might be due to a ceiling effect in the test as to its focus on fundamentals.

\subsection{RQ2 - Students' perception and engagement}

\begin{figure}[h]
  \centering
  \vspace{-5pt}
  \includegraphics[width=0.6\linewidth]{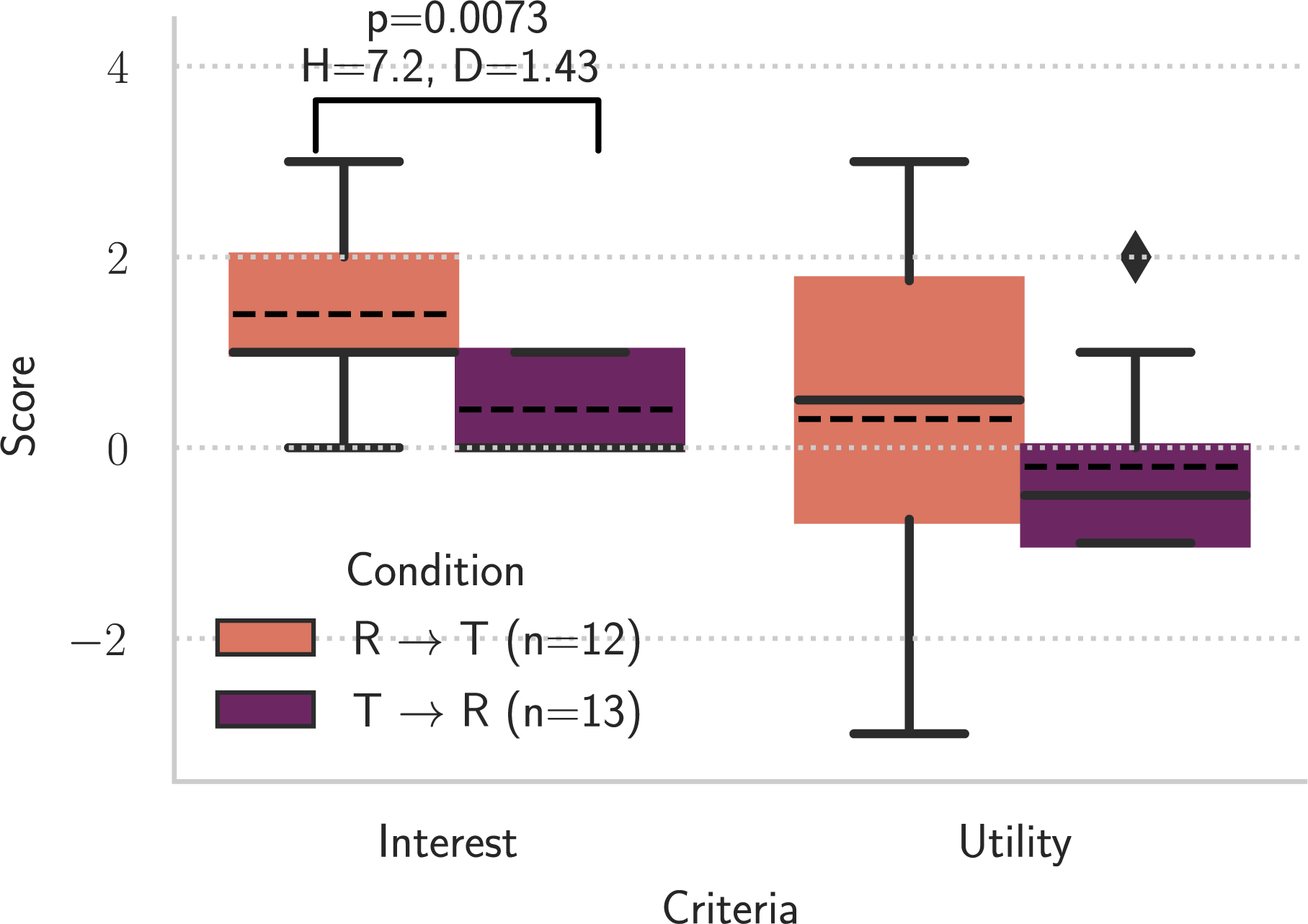}
  \vspace{-10pt}
  \caption{Students' interest and utility assessment of the ER-based lecture. Significant Kruskal Wallis tests are indicated with p-value, H statistic and Cohen's effect size (D).}
  \vspace{-10pt}
  \label{fig:perception_robotics_theory}
\end{figure}

Fig. \ref{fig:perception_robotics_theory} reports the students' perception of the ER-based theoretical lecture, in terms of interest and utility, measured via the \textit{Robotics theory} survey. Students in the RT condition, who started with the ER-based lecture, perceived it as significantly more interesting (Kruskal Wallis p=0.0073, H=7.3, D=1.43), and, although not significantly, also more useful than those in the TR condition, who did it after the traditional lecture. This finding supports the results of RQ1 in suggesting that the two types of lectures are equally valid in transmitting relevant knowledge to the students, with those in condition TR thus finding the ER-based lecture of little interest and utility.
Conversely, one of the teachers expressed a preference towards the TR condition, described as closer to the current practice and both manifested interest towards ER-based mathematics.
 
\begin{figure}[t]
  \centering
  \vspace{-8pt}
  \includegraphics[width=0.6\linewidth]{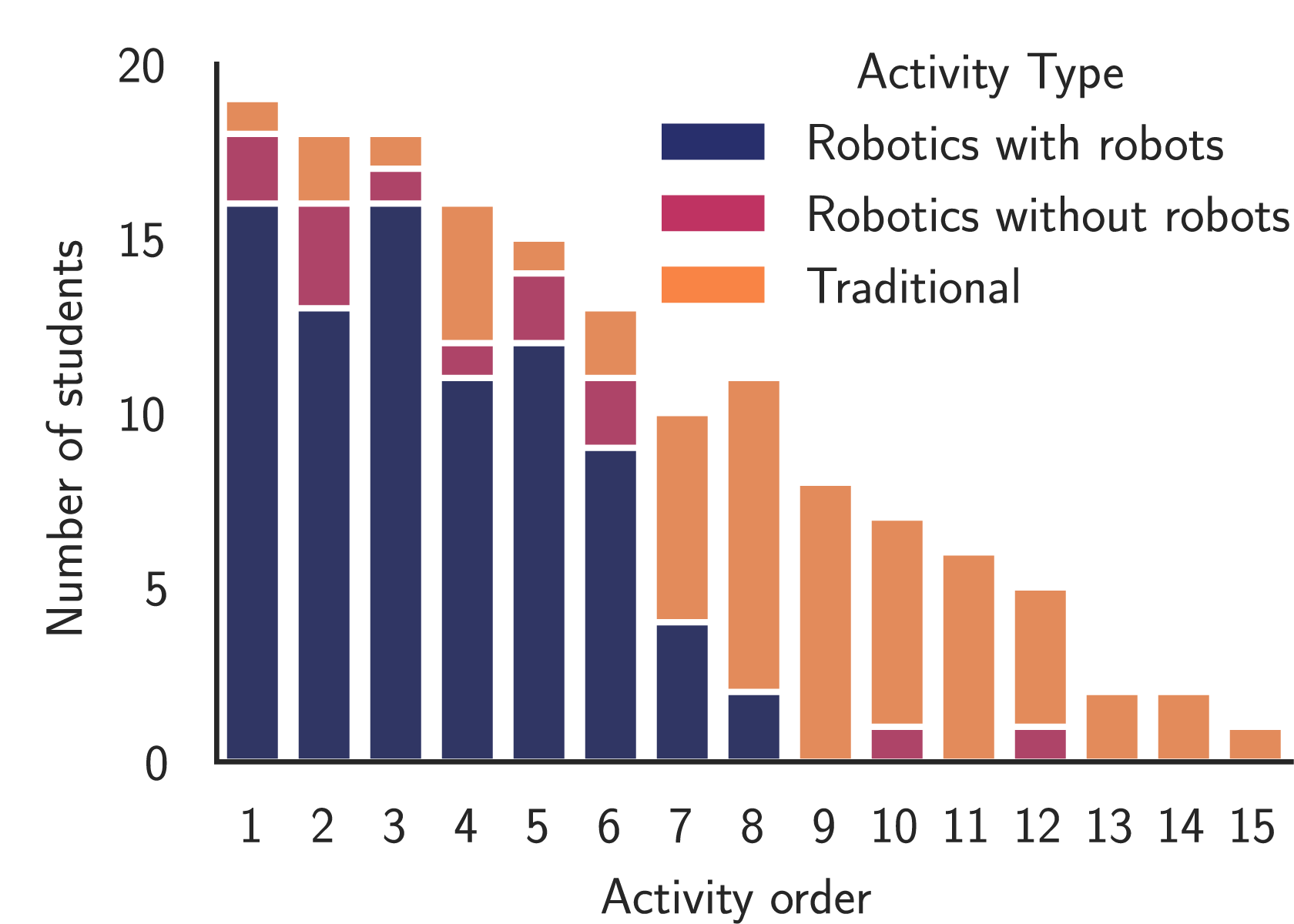}
  \vspace{-10pt}
  \caption{Order in which the exercises were conducted. 
  }
  \vspace{-5pt}
  \label{fig:activity_order}
\end{figure}

\begin{figure}[t]
  \centering
  \vspace{-5pt}
  \includegraphics[width=0.6\linewidth]{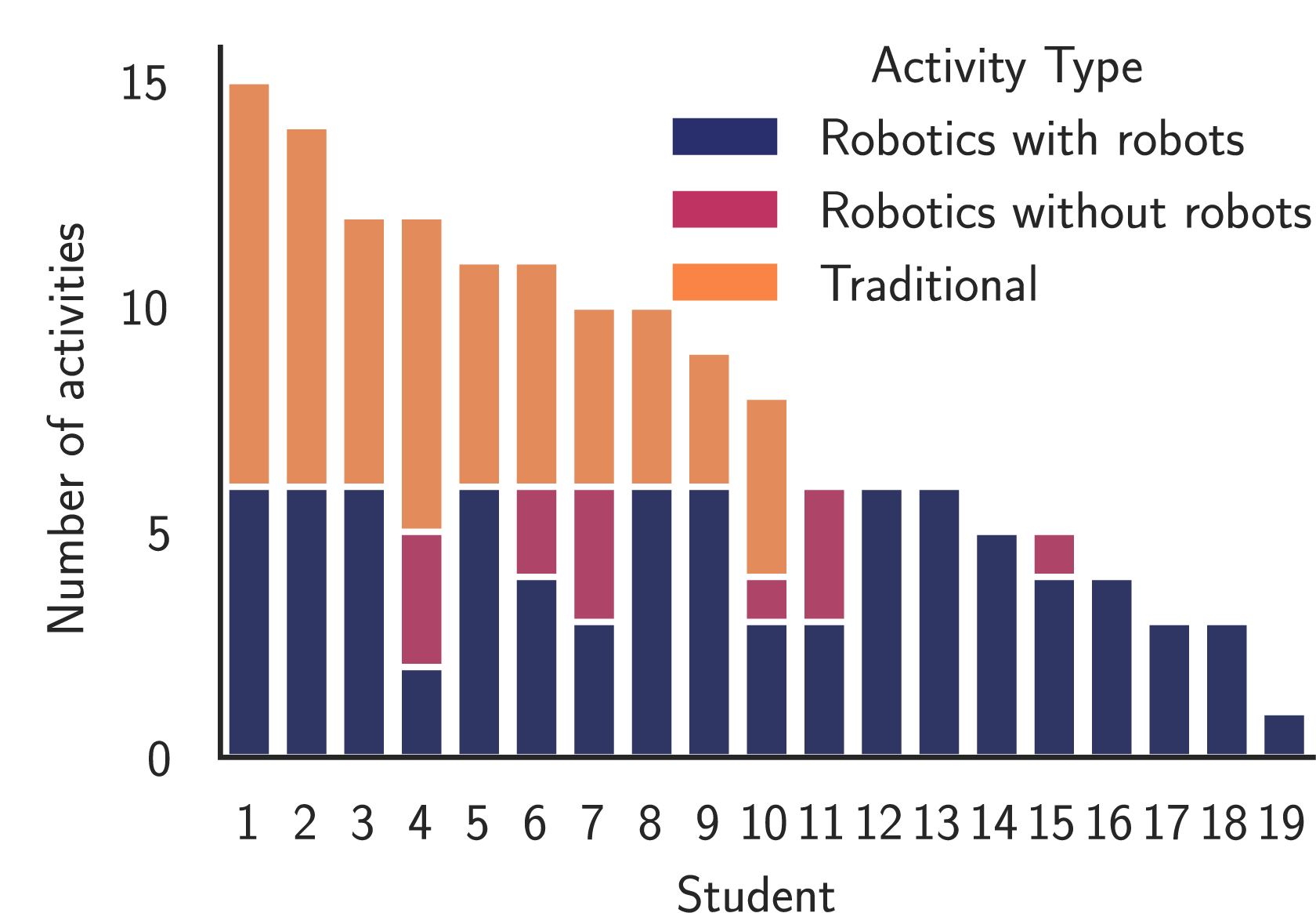}
  \vspace{-10pt}
  \caption{Number of exercises conducted by each student.}
  \vspace{-10pt}
  \label{fig:activity_students}
\end{figure}

To investigate students' engagement with the ER-based exercises proposed during the exercise sessions, we extracted from their \textit{Roadmap} documents the order in which the activities were done by students (shown in Fig. \ref{fig:activity_order}) and the number of activities each student did individually (see Fig. \ref{fig:activity_students}). In both analyses, we distinguish between traditional exercises, ER-based exercises solved without using the robot (``robotics without robot'') and ER-based exercises solved using the robot (``robotics with robot'').
Fig. \ref{fig:activity_order} shows that most students started with ``robotics with robot'' exercises and finished with the traditional exercises, with few ``robotics without robot'' exercises being conducted overall.
Since only 6 ER-based exercises were designed, the figure suggests that most students not only did many of them, but also did them in block, before transitioning to traditional ones.
Indeed, Fig. \ref{fig:activity_students} shows that all the students conducted at least one ER-based exercise with the robot (``robotics with robots'', $\mu = 4.15 \pm 1.53$), with only 6 students engaging in ``robotics without robot'' exercises. Moreover, since the students who did the largest number of exercises also did all of the ``robotics with robot'' exercises, it would seem that the time spent on ER-based exercises was not detrimental for their overall engagement with the exercises.
It is important to note that 5 sessions were allocated to exercises (as per curriculum), and several students missed one or more of them\footnote{also due to COVID-19 regulations concerning in-presence and distance learning.}, leading to lower-than-average number of exercises conducted within the allotted time.

\begin{figure}[t]
  \centering
  \vspace{-5pt}
  \includegraphics[width=0.6\linewidth]{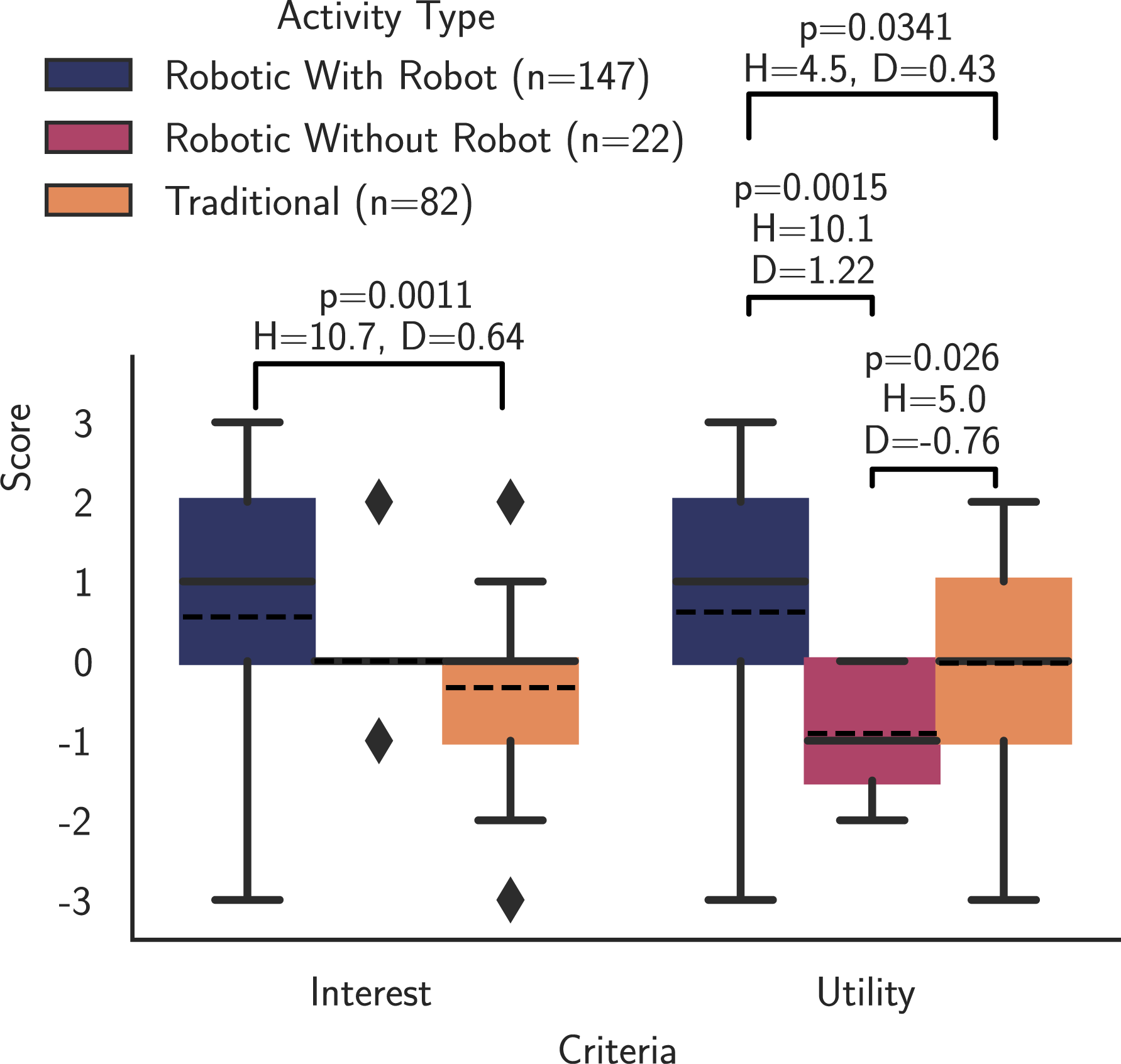}
  \vspace{-10pt}
  \caption{Students' interest and utility assessment of the ER-based exercises. Significant Kruskal Wallis tests are reported with p-value, H statistic and Cohen’s effect size (D).}
  \label{fig:perception_of_the_activities}
  \vspace{-10pt}
\end{figure}

Fig. \ref{fig:perception_of_the_activities} reports the students' perception of the ER-based exercises, in terms of interest and utility, comparing the ratings of ``robotics with robot'', ``robotics without robot'' and ``traditional'' exercises.
``Robotics with robot'' exercises were perceived as interesting ($\mu = 0.5 \pm 1.5$)), while ``robotics without robot'' and traditional ones were rated more negatively ($\mu = 0.0 \pm 0.7$,  $\mu = -0.33 \pm 1.2$ respectively). This contributes to a significant difference in interest between the ER-based exercises done with the robot and the traditional ones (Kruskal Wallis test $p=0.0011$, $H=10.7$, $D=0.64$). 
Similarly, ``robotics with robot'' exercises are perceived as useful ($\mu = 0.6 \pm 1.6$) and significantly more so than the traditional ones (mediocre utility, $\mu = -0.9 \pm 0.8$) and ``robotics without robot'' ones ($\mu = -0.02 \pm 1.4$). ``Robotics without robot'' exercises are not only judged less favourably that their ``with robot'' counterparts, but also perceived as less useful than traditional activities, which suggests that the role of the robot in the ER-based exercises was meaningful, allowing for the creation of novel exercises relying on different modalities to convey and verify a same content. 

\begin{figure}[ht]
  \centering
  \vspace{-10pt}
  \includegraphics[width=0.7\linewidth]{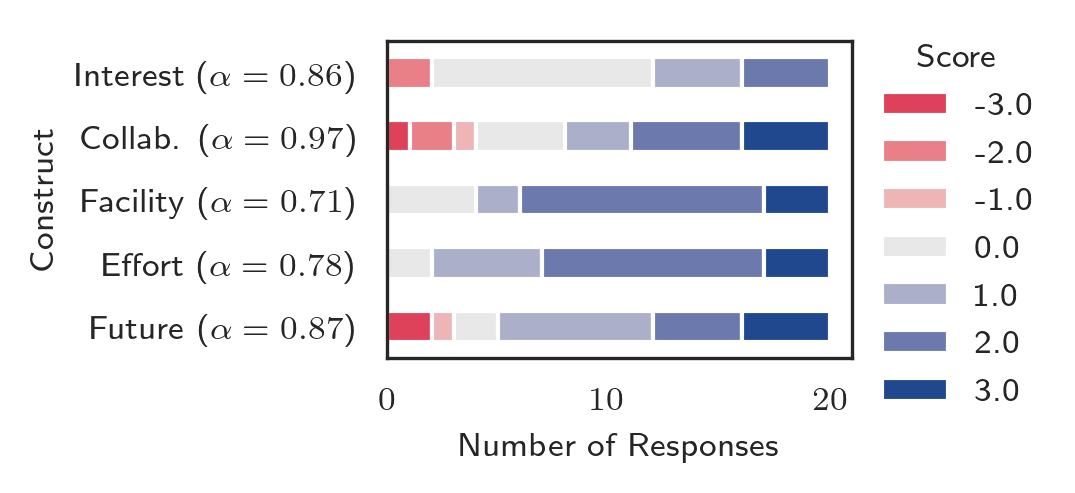}
  \vspace{-15pt}
  \caption{Students' perception of the ER-based content. 
  Cronbach's $\alpha$ internal consistency of the scale is calculated for each construct and shown in parenthesis.
  }
  \vspace{-15pt}
  \label{fig:overall_perception}
\end{figure}

In the \textit{perception survey}, students evaluated the ER-based content from the perspectives of interest, collaboration, facility, effort put in, and future interest (see Fig. \ref{fig:overall_perception}). Students had a globally positive opinion of the ER-based content and reported high interest, facility and effort with respect to integrating robotics in the geometry lecture. Although collaboration and future interest obtained slightly lower scores than the others constructs, the results remained globally positive ($\mu =0.9 \pm 1.9$ and $\mu = 1.0 \pm 1.6$ respectively). 
However, some students observed that the robot ``lacked a bit of precision for the constructions''. Indeed, the robot's motion accuracy was not-always meeting the requirements of the application, which, together with some connectivity issues (causing the robot to skip certain instructions), caused some frustration for the students.

\subsection{RQ3 - Effect of prior appreciation for maths}

\begin{figure}[ht]
  \centering
  \includegraphics[width=0.7\linewidth]{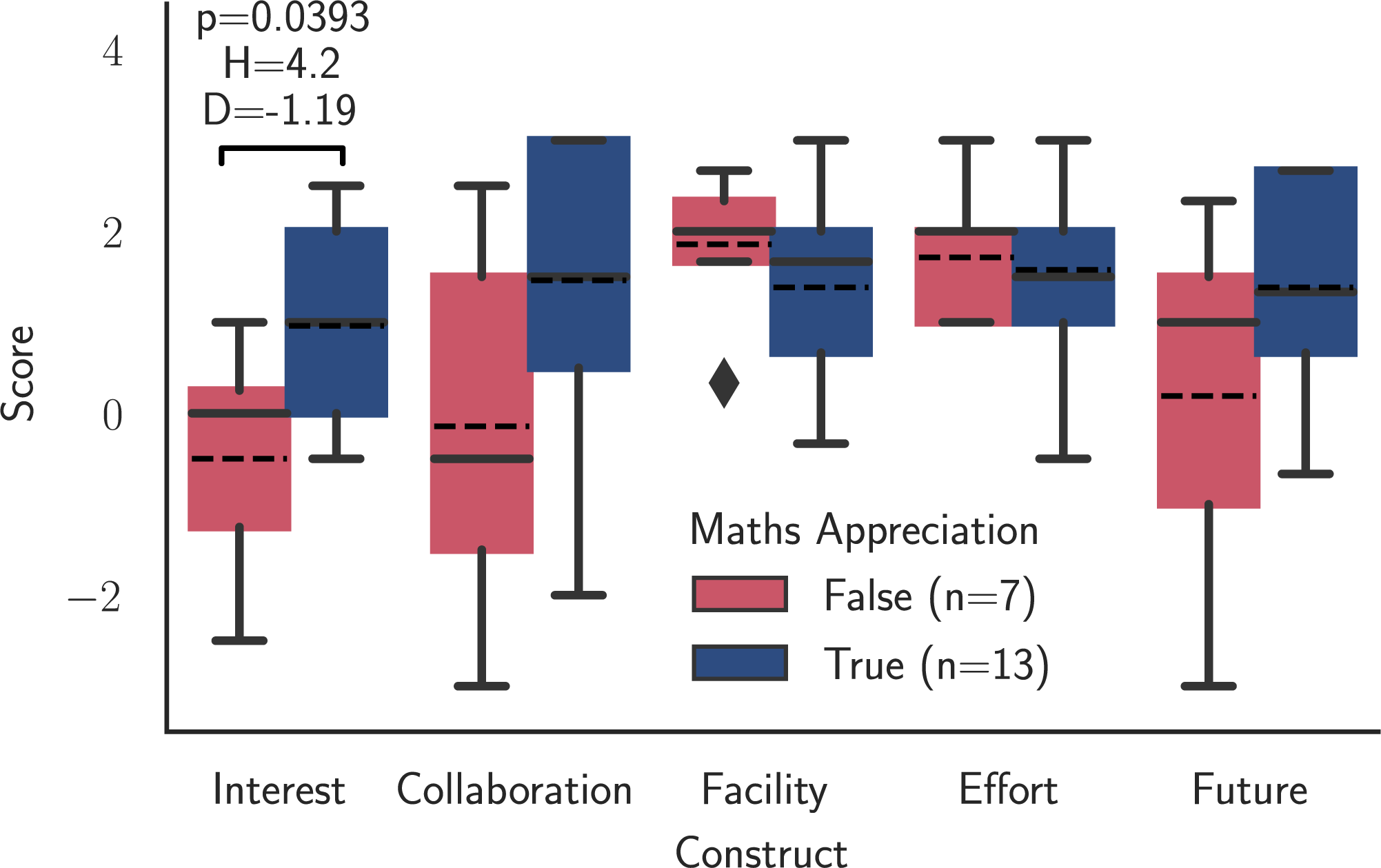}
  \vspace{-5pt}
  \caption{Students' perception of the ER-based content depending on prior maths appreciation. Significant Kruskal Wallis tests are reported with p-value, H statistic and Cohen’s effect size (D).}
  \vspace{-15pt}
  \label{fig:math_appreciation}
\end{figure}

The level of engagement in the exercise sessions (extracted from the \textit{Roadmap}) was compared between the students who liked and those who did not like mathematics, to verify the effect of this variable on our observed ones. No significant differences were found between the students in terms of number of ``robotics with robot'', ``robotics without robot'', nor traditional exercises they engaged in (Kruskal Wallis test fails to reject $H_0$, $p>0.05$). 
Similarly, no significant differences were found between the students in terms of knowledge acquired, both at the end of the theoretical lectures and at the end of the unit (Test 2 and Test 3, see Section \ref{RQ1}), although it is possible that they were affected by a ceiling effect. 
Fig. \ref{fig:math_appreciation} compares the responses given to the final \textit{Perception} survey by the students who liked and those who did not like mathematics. 
No significant differences were found between these groups (Kruskal Wallis test fails to reject $H_0$, $p>0.05$) for collaboration, facility, effort and future interest. Conversely, students who don't like math perceive the ER-based content as less interesting than their classmates who appreciate the discipline, a finding that can be read as the proof that students were not fooled by the novelty introduced by the robot and perceived the ER-based content as geometry content. 
These findings (albeit limited in validity by the low number of students not liking mathematics) would seem to suggest that a hybrid between traditional and ER-based content has the potential to engage students who generally don't like mathematics as much as those who do. As a consequence, these students might possibly improve their competence in and appreciation of mathematics. Indeed, it would be interesting to verify this hypothesis in a longitudinal study, concurrently tracking students' perception of mathematics and of Robotics. 

\section{Discussion and Conclusion}

This article investigates modes and benefits of the introduction of Educational Robotics in the formal curriculum of secondary school mathematics, specifically focusing on the 7-hours long learning unit about planar geometric figures, that students in Swiss schools address at grade 11 (15 y.o.).
The study, involving 26 students from two classes, started with the preparation of a 90-minute long ER-based theoretical lecture and 6 ER-based exercises, validated by 5 experts to ensure their alignment with the learning objectives of the unit and state-of-art teaching practices. 
Rather than designing a fully robotics-based geometry course, our objective was to include a limited set of activities where the robot had an added value. Indeed, robotics should be considered as a tool, an extension to traditional paper-based methods, but not a total replacement.
We specifically investigated (RQ1) the role of ER-based theoretical lectures, with respect to traditional ones; (RQ2) students' perception of and engagement in the proposed ER-based lecture and exercises; (RQ3) whether the findings of RQ1 and RQ2 differ according to students' prior appreciation for mathematics. 

To investigate RQ1, half of the students started with the ER-based theoretical lecture and the other half with the traditional lecture, and then switched. Results showed that both groups of students, after their first lecture, reached a similar (and similarly high) level of competence, suggesting that ER, more often associated with exercises than theoretical lectures, can be as effective as traditional means to introduce abstract concepts. 
While the order in which students received the two types of theoretical lecture had no impact on their learning, likely due to the double exposure, the students evaluated the ER-based lecture more positively when done first. As different hypotheses can be made as to what caused these differences in perception, future studies should be envisioned to further and specifically investigate this topic. 
Students showed a generally positive perception of the ER-based content (RQ2), both through their assessment in terms of interest, utility, collaboration, facility, effort and interest for future integration and their behaviour during the exercise sessions. 
Indeed, most students started with, and engaged in all of, the ER-based exercises, despite the limitations of the platform sometimes frustrating their efforts.
We hypothesise that a key reason for this success, to be verified in future studies, is that ER allows the student to be actively engaged in the exercise and provides immediate visual feedback that helps avoid the fear of errors and judgement.
Lastly, results were generally similar between students who had declared liking mathematics, and those who had declared not to (RQ3), encouraging us to investigate in a future long-term study whether ER could possibly have a positive impact on students' competence in, and perception of, mathematics. 

In more general terms, the findings highlight the importance of considering, and doing so as early as possible, the alignment between the learning outcomes and the robotics artefacts \cite{Giang_281683}, the requirements posed by the classroom context \cite{shahmoradi2020teachers}, as well as discipline-specific ones (e.g., the need for precise localisation required by geometry). 
While our preliminary findings should be verified in broader and longer studies, the most important result of this study may be its standing as proof not only that ER can be introduced in formal education, but also, thanks to the increasing efforts to train teachers \cite{el-hamamsy_computer_2021, el-hamamsy_symbiotic_2021}, that this can be done in a way that allows research and teaching practices to coexist and mutually benefit from their interplay within a context of translational research. 

\section{ACKNOWLEDGMENTS}

We would like to thank our colleagues (E.B, M.S), the members of the school (D.S., A.L.), teachers (M.H., S.K., B.J-D., Y.G.) and students who helped set afoot the experiments. 

\ifdefined\USEACM
\bibliographystyle{ACM-Reference-Format}
\else 
\bibliographystyle{splncs04}
\fi
\bibliography{biblio} 

\begin{thebibliography}{10}
\providecommand{\url}[1]{\texttt{#1}}
\providecommand{\urlprefix}{URL }
\providecommand{\doi}[1]{https://doi.org/#1}

\bibitem{alimisis2013educational}
Alimisis, D.: Educational robotics: Open questions and new challenges. Themes
  in Science and Technology Education  \textbf{6}(1),  63--71 (2013)

\bibitem{Azevedo_1995}
Azevedo, R., Bernard, R.M.: A meta-analysis of the effects of feedback in
  computer-based instruction. Journal of Educational Computing Research
  \textbf{13}(2),  111--127 (1995)

\bibitem{benitti_how_2017}
Benitti, F.B.V., Spolaôr, N.: How {Have} {Robots} {Supported} {STEM}
  {Teaching}? In: Khine, M.S. (ed.) Robotics in {STEM} {Education}:
  {Redesigning} the {Learning} {Experience}, pp. 103--129. Springer, Cham
  (2017)

\bibitem{bers_teaching_2005}
Bers, M.U., Portsmore, M.: Teaching {Partnerships}: {Early} {Childhood} and
  {Engineering} {Students} {Teaching} {Math} and {Science} {Through}
  {Robotics}. Journal of Science Education and Technology  \textbf{14}(1),
  59--73 (Mar 2005)

\bibitem{chevalier_fostering_2020}
Chevalier, M., Giang, C., Piatti, A., Mondada, F.: Fostering computational
  thinking through educational robotics: a model for creative computational
  problem solving. IJ STEM Ed  \textbf{7}(1), ~39 (Dec 2020)

\bibitem{chevalier_pedagogical_2016}
Chevalier, M., Riedo, F., Mondada, F.: Pedagogical {Uses} of {Thymio} {II}:
  {How} {Do} {Teachers} {Perceive} {Educational} {Robots} in {Formal}
  {Education}? IEEE Robotics \& Automation Magazine (RAM)  \textbf{23(2)},
  16--23 (2016)

\bibitem{cronbach2004my}
Cronbach, L.J., Shavelson, R.J.: My current thoughts on coefficient alpha and
  successor procedures. Educational and psychological measurement
  \textbf{64}(3),  391--418 (2004)

\bibitem{Eguchi}
Eguchi, A.: Educational robotics theories and practice: Tips for how to do it
  right. In: Robots in K-12 education: A new technology for learning, pp.
  1--30. IGI Global (2012)

\bibitem{el-hamamsy_symbiotic_2021}
El-Hamamsy, L., Bruno, B., Chessel-Lazzarotto, F., Chevalier, M., Roy, D.,
  Zufferey, J.D., Mondada, F.: The symbiotic relationship between educational
  robotics and computer science in formal education. Educ Inf Technol  (Apr
  2021)

\bibitem{el-hamamsy_computer_2021}
El-Hamamsy, L., Chessel-Lazzarotto, F., Bruno, B., Roy, D., Cahlikova, T.,
  Chevalier, M., Parriaux, G., Pellet, J.P., Lanarès, J., Zufferey, J.D.,
  Mondada, F.: A computer science and robotics integration model for primary
  school: evaluation of a large-scale in-service {K}-4 teacher-training
  program. Educ Inf Technol  \textbf{26}(3),  2445--2475 (May 2021)

\bibitem{ollero_methodology_2018}
Ferrarelli, P., Lapucci, T., Iocchi, L.: Methodology and results on teaching
  maths using mobile robots. In: Iberian Robotics conference, pp. 394--406
  (2017)

\bibitem{foerster_integrating_2016}
Foerster, K.T.: Integrating programming into the mathematics curriculum:
  Combining scratch and geometry in grades 6 and 7. In: Proceedings of the 17th
  annual conference on information technology education. pp. 91--96 (2016)

\bibitem{fowler1995improving}
Fowler~Jr, F.J., Fowler, F.J.: Improving survey questions: Design and
  evaluation. Sage (1995)

\bibitem{Giang_281683}
Giang, C.: Towards the alignment of educational robotics learning systems with
  classroom activities p.~176 (2020)

\bibitem{Pair_programming}
Hanks, B., Fitzgerald, S., McCauley, R., Murphy, L., Zander, C.: Pair
  programming in education: a literature review. Computer Science Education
  \textbf{21}(2),  135--173 (2011)

\bibitem{Iskrenovic_Momcilovic_Olivera}
Iskrenovic-Momcilovic, O.: Improving geometry teaching with scratch.
  International Electronic Journal of Mathematics Education  \textbf{15} (02
  2020)

\bibitem{jung_systematic_2018}
Jung, S., Won, E.s.: Systematic {Review} of {Research} {Trends} in {Robotics}
  {Education} for {Young} {Children}. Sustainability  \textbf{10}(4), ~905 (Mar
  2018)

\bibitem{karim2015review}
Karim, M.E., Lemaignan, S., Mondada, F.: A review: Can robots reshape k-12 stem
  education? In: 2015 IEEE international workshop on Advanced robotics and its
  social impacts (ARSO). pp.~1--8. IEEE (2015)

\bibitem{Hee_2020}
Kim, H.J.: Concreteness fading strategy: A promising and sustainable
  instructional model in mathematics classrooms. Sustainability  \textbf{12},
  ~2211 (03 2020)

\bibitem{King_2006}
King, W., He, J.: A meta-analysis of the technology acceptance model.
  Information \& Management  \textbf{43},  740--755 (09 2006)

\bibitem{leoste_impact_2019}
Leoste, J., Heidmets, M.: The impact of educational robots as learning tools on
  mathematics learning outcomes in basic education. In: Digital Turn in
  Schools—Research, Policy, Practice. pp. 203--217. Springer (2019)

\bibitem{miller2016robotics}
Miller, D.P., Nourbakhsh, I.: Robotics for education. In: Springer handbook of
  robotics, pp. 2115--2134. Springer (2016)

\bibitem{mondada_bringing_2017}
Mondada, F., Bonani, M., Riedo, F., Briod, M., Pereyre, L., Retornaz, P.,
  Magnenat, S.: Bringing {Robotics} to {Formal} {Education}: {The} {Thymio}
  {Open}-{Source} {Hardware} {Robot}. IEEE Robotics Automation Magazine
  \textbf{24}(1),  77--85 (Mar 2017)

\bibitem{papert_mindstorms:_1980}
Papert, S.: Mindstorms: children, computers, and powerful ideas. Basic Books,
  New York (1980)

\bibitem{resnick2009scratch}
Resnick, M., Maloney, J., Monroy-Hern{\'a}ndez, A., Rusk, N., Eastmond, E.,
  Brennan, K., Millner, A., Rosenbaum, E., Silver, J., Silverman, B., et~al.:
  Scratch: programming for all. Communications of the ACM  \textbf{52}(11),
  60--67 (2009)

\bibitem{Rogers2004BringingET}
Rogers, C., Portsmore, M.D.: Bringing engineering to elementary school. Journal
  of STEM Education: Innovations and Research  \textbf{5},  17--28 (2004)

\bibitem{ryan2000self}
Ryan, R.M., Deci, E.L.: Self-determination theory and the facilitation of
  intrinsic motivation, social development, and well-being. American
  psychologist  \textbf{55}(1), ~68 (2000)

\bibitem{shahmoradi2020teachers}
Shahmoradi, S., Kothiyal, A., Olsen, J.K., Bruno, B., Dillenbourg, P.: What
  teachers need for orchestrating robotic classrooms. In: European Conference
  on Technology Enhanced Learning. pp. 87--101. Springer (2020)

\bibitem{silk2010designing}
Silk, E.M., Higashi, R., Shoop, R., Schunn, C.D.: Designing technology
  activities that teach mathematics. The Technology Teacher  \textbf{69}(4),
  21--27 (2010)

\bibitem{zhong_systematic_2020}
Zhong, B., Xia, L.: A {Systematic} {Review} on {Exploring} the {Potential} of
  {Educational} {Robotics} in {Mathematics} {Education}. International Journal
  of Science and Mathematics Education  \textbf{18}(1),  79--101 (Jan 2020)

\end{thebibliography}

\end{document}